\pdfoutput=1
\documentclass[11pt]{article}

\usepackage[preprint]{acl}

\usepackage{times}
\usepackage{latexsym}

\usepackage{multirow}

\usepackage[T1]{fontenc}
\usepackage[utf8]{inputenc}

\usepackage{microtype}

\usepackage{inconsolata}

\usepackage{graphicx}

\title{The Challenge of Achieving Attributability in Multilingual Table-to-Text Generation with Question-Answer Blueprints}

\author{Aden Haussmann \\
  The University of Edinburgh \\
  \texttt{adenrhaussmann@gmail.com}}

\begin{document}
\maketitle
\begin{abstract}
Multilingual Natural Language Generation (NLG) is challenging due to the lack of training data for low-resource languages. However, some low-resource languages have up to tens of millions of speakers globally, making it important to improve NLG tools for them. Table-to-Text NLG is an excellent measure of models’ reasoning abilities but is very challenging in the multilingual setting. System outputs are often not attributable, or faithful, to the data in the source table. Intermediate planning techniques like Question-Answer (QA) blueprints have been shown to improve attributability on summarisation tasks. This work explores whether QA blueprints make multilingual Table-to-Text outputs more attributable to the input tables. This paper extends the challenging multilingual Table-to-Text dataset, TaTA, which includes African languages, with QA blueprints. Sequence-to-sequence (seq2seq) language models are then finetuned on this dataset, with and without blueprints. Results show that QA blueprints improve performance for models finetuned and evaluated only on English examples, but do not demonstrate gains in the multilingual setting. This is due to inaccuracies in machine translating the blueprints from English into target languages when generating the training data, and models failing to rely closely on the blueprints they generate. An in-depth analysis is conducted on why this is challenging.
\end{abstract}

\section{Introduction}

The majority of NLP research, models and datasets focus on English \citep{ruder2022statemultilingualai}. Yet there are many widely-spoken languages which are severely underrepresented (low-resource languages, or LRLs). For example, Igbo is a language found predominantly in Nigeria that is considered low-resource despite being spoken by around 44 million people.\footnote{\url{https://celt.indiana.edu/portal/Igbo/index.html}} It is important that more research is done to investigate techniques which make NLP work better for a more diverse set of languages.

\begin{figure}[h!]
    \centering
    \includegraphics[width=\columnwidth]{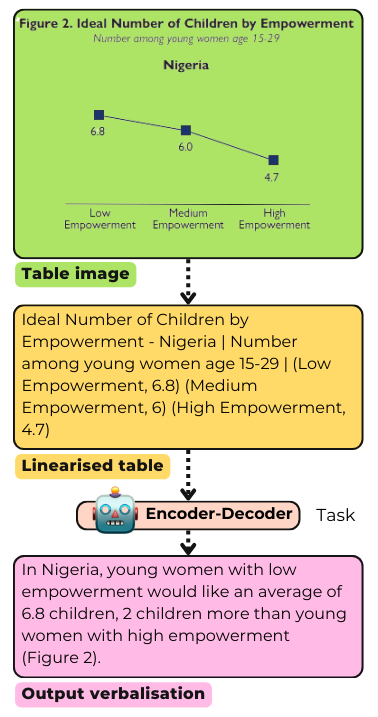}
    \caption{An example of a chart image, its linearised form (created by human annotators), and an output verbalisation (produced by a seq2seq neural network).}
    \label{fig:linearisation}
\end{figure}

The task explored in this paper is to generate \textbf{fluent} and \textbf{accurate} descriptions of the data in the input table. A seq2seq model takes the linearised table (in textual form, not the image of the table itself) as input, and the goal is to output a verbalisation, like in \autoref{fig:linearisation}.

In this paper, conditional generation with Question-Answer blueprints, a technique which has proven effective for improving the faithfulness on the summarisation task, is applied to the task of Table-to-Text generation for the first time. Further, it is applied to the challenging multilingual dataset, T\textsc{a}TA, or \textit{Table-to-Text in African languages} \citep{tata}.

This paper evaluates whether finetuning a seq2seq neural language model with QA blueprints improves the attributability (faithfulness to the source table) of output verbalisations.

The results show that QA blueprints do increase attributability of outputs for models finetuned and evaluated on only English data. However, in the multilingual setting the technique is less effective. The challenge is twofold; inaccuracies in machine translating blueprints generated in English into target languages makes the training dataset less than perfect, baking in a fundamental disadvantage before training even begins. Furthermore, models struggle to produce verbalisations which rely heavily on their blueprints. This is analysed in detail in \autoref{sec:results}.

This paper tests the performance of a range of metrics on the task, including \textsc{chr}F \citep{chrf}, BLEU \citep{bleu} and F\textsc{act}KB \citep{feng2023factkb}, and retrains the T\textsc{a}TA-specific learned metric, S\textsc{t}ATA, or \textit{Statistical Assessment of Table-to-Text in African languages}, introduced by \citet{tata} and makes it publicly available. See \autoref{sec:stata} for more details.

\section{Related Work}

Neural language models tend to hallucinate and repeat themselves, and they struggle to remain faithful to input data and identify important information. Solving these problems is challenging because deep neural networks are opaque by nature, making it difficult to understand their reasoning and find the root of errors \citep{qa-generation}.

\begin{figure*}[t]
    \centering
    \includegraphics[width=\textwidth]{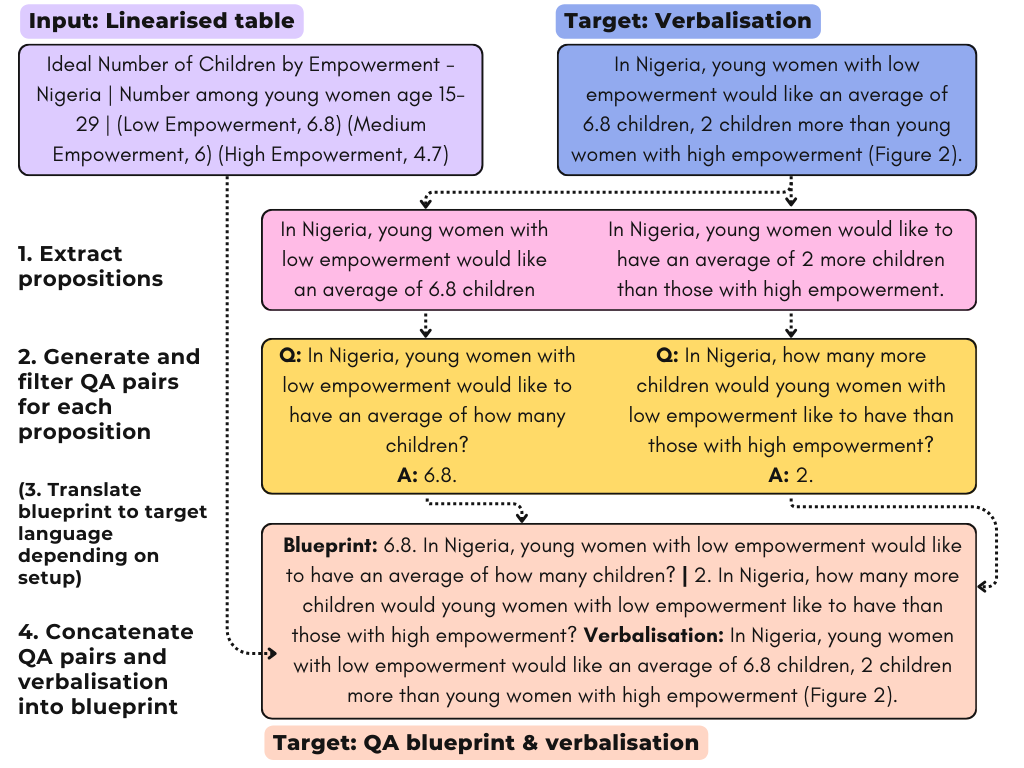}
    \caption{The process of generating QA blueprints.}
    \label{fig:blueprint-creation}
\end{figure*}

Numerous techniques have been applied in attempts to reduce hallucination and make outputs more faithful and attributable to the source document \citep{qa-generation}. One family of such techniques is centred around content selection and planning, whereby the model is trained to identify and extract relevant information from the input, and generate an ``intermediate plan'' of \textit{what to say} (selection) and \textit{in what order} (planning) before generating the actual output, which is conditioned on the plan.

One idea is to build the intermediate plans from question and answer (QA) pairs, where the questions and answers are generated from the target summaries. This has been shown to allow more control of a model's output, and more explanation of a model's features than entity chains \citep{qa-generation}.

Content selection and planning has also proved effective in improving Table-to-Text outputs \citep{content-planning}. In 2019 such a model achieved state-of-the-art BLEU scores on the RotoWire \citep{wiseman-etal-2017-challenges} and ToTTo \citep{totto} datasets. This work used an LSTM-based encoder-decoder architecture with an attention mechanism and has since been surpassed by simply finetuning T5\textsubscript{XL} \citep{t5}. This showed that the modern text-to-text pretraining-finetuning paradigm employed by models such as mT5 works very well for Table-to-Text tasks \citep{kale-rastogi-2020-text}.

mT5 \citep{mt5}, the multilingual variant of T5, is used in this paper.

\section{QA Blueprint Creation}

A similar process for creating the QA blueprints is followed to that of \citet{qa-generation}, but with some differences. The process is summarised in Figure \ref{fig:blueprint-creation}.

\subsection{Extracting propositions}

The following is done for each English reference (which are single sentences):

First, propositions are extracted from the reference. FLAN-T5-Large \citep{chung2022scaling} finetuned\footnote{\url{https://huggingface.co/chentong00/propositionizer-wiki-flan-t5-large}} for ``propositionizing'' sentences by \citet{chen2023dense} is used generate a minimal sentence for each proposition in the reference.

For example, for the following sentence:

\begin{quote}
    \textit{In Nigeria, young women with low empowerment would like an average of 6.8 children, 2 children more than young women with high empowerment (Figure 2).}
\end{quote}

the following propositions are extracted:

\begin{itemize}
    \item \textit{In Nigeria, young women with low empowerment would like a average of 6.8 children.}
    \item \textit{In Nigeria, young women with low empowerment would like to have an average of 2 more children than those with high empowerment.}
\end{itemize}

\subsection{Generating \& Heuristically Selecting QA Pairs}

The QA pairs are generated directly from each proposition, instead of the reference itself. This guarantees that there will not be overlapping or redundant information, i.e. each QA pair expresses a different fact, assuming there is no information overlap between propositions.

For each proposition, 5 diverse QA pairs are generated using T5-Large finetuned\footnote{\url{https://huggingface.co/potsawee/t5-large-generation-squad-QuestionAnswer}} on SQuAD \citep{rajpurkar-etal-2016-squad} for QA generation by \citet{mqag}.

Next, QA pairs where the question does not end with a question mark, or the question or answer are just empty strings, are dropped.

The goal of the QA blueprints is to focus the model's output on factual data which is attributable to the input table. If numbers are hallucinated at this point, this error will propagate into the final output verbalisation and defeat the blueprint's purpose. Therefore, pairs containing a hallucinated number, that is not present in the source table, are dropped.

QA pairs where the answer is fully contained within the question are dropped.

For any duplicated answers, the QA pair where the question has the greatest lexical similarity with the proposition (calculated via a word-level F1 score between the question and reference), is chosen. The other is dropped.

Finally, the QA pair with the greatest lexical similarity with the reference is selected. For each of the propositions in the above example respectively, these are:

\begin{itemize}
    \item \textit{``Question: In Nigeria, young women with low empowerment would like an average of how many children? Answer: 6.8.''}
    \item \textit{``Question: In Nigeria, how many more children would young women with low empowerment like to have than those with high empowerment? Answer: 2.''}
\end{itemize}

On average, a QA blueprint generated via this strategy contains two or three QA pairs.

\subsection{Building the Blueprint}

The chosen QA pairs are then concatenated as follows: Let \textit{a} denote an answer, and \textit{q} denote a question. The blueprint \textit{b} takes the form $a_1$;$q_1$;...;$a_n$;$q_n$. Although it is less natural, answers come before questions following the procedure by \citet{qa-generation} that yielded superior results. Answers and questions are separated by a full stop ``.'', and QA pairs are separated by a pipe ``$|$''.

Special tokens \textit{``Blueprint:''} and \textit{``Verbalisation:''} are prefixed to the blueprint \textit{b} and verbalisation \textit{v} respectively, and these are concatenated to form the references.

In this setup, the encoder-decoder model takes the linearised table \textit{t} as input and learns to predict the blueprint $b$ as $p(b|t)$, then generate the output $v$ as $p(v|t, b)$. It is noted by \citet{qa-generation} that this relies on the blueprints being correct, and acknowledged that errors at this stage will propagate down the pipeline and affect the final generated verbalisation.

\subsection{Making Blueprints Multilingual} \label{sec:making-mult}

Since the goal is to produce multilingual generations, two blueprint setups are created. The first just uses the English blueprints. So, when the model generates output $v$ as $p(v|t, b)$, $b$ is always in English and $v$ is in the target language. The second still generates the blueprints in English but translates them to the target language. So, both $b$ and $v$ are in the target language. For the English blueprint dataset only, a language tag, for example \textit{``(Yor\`ub\'a)''} is also inserted after \textit{``Verbalisation''} to help indicate to the model what the target language for the verbalisation is.

Machine translating QA pairs from English to the other languages will propagate small errors in translation further down the pipeline and hinder the effectiveness of the blueprints. To evaluate this risk, an analysis of translation quality is done on the training set. Since the dataset is parallel (i.e., each sample appears in every language), there is ample data to test the quality of machine translations in this specific context.

Each English sample's reference is translated, using Google Translate,\footnote{\url{https://pypi.org/project/googletrans/}} into the seven other languages which appear in the training set, and these translations are compared to the corresponding samples written in the target languages with automatic MT metrics. The results are recorded in \autoref{tab:translations}.

\begin{table*}[t]
\centering
\begin{tabular}{cccccccc}
\hline
              & \textbf{Hausa} & \textbf{Igbo} & \textbf{Swahili} & \textbf{Yor\`ub\'a} & \textbf{French} & \textbf{Arabic} & \textbf{Portuguese} \\ \hline
\textbf{\textsc{chr}F} & 0.53           & 0.56          & 0.68             & 0.19            & 0.80            & 0.57            & 0.66                \\
\textbf{BLEU} & 0.30           & 0.37          & 0.47             & 0.06            & 0.68            & 0.33            & 0.47                \\ \hline
\end{tabular}
\caption{\textsc{chr}F and BLEU scores for Google Translate-powered translations of English samples in T\textsc{a}TA compared to corresponding samples in the target languages.}
\label{tab:translations}
\end{table*}

Translation quality for the more broadly spoken languages such as French and Portuguese is high, with lower quality for the African languages. This is to be expected, as mT5 will have seen more English and French samples during its training than, say, Igbo. Swahili enjoys the best translations of the African languages, achieving higher scores than even Arabic. All languages can be translated with reasonable quality, with the exception of Yor\`ub\'a, which achieves the lowest scores by a large margin.

See \autoref{sec:appendix-translations} for an example translated blueprint in Swahili.

\subsection{Repetition Penalty}

It is observed that the finetuned models are all highly susceptible to very repetitive outputs, especially non-English languages, and the effect is worsened when using blueprints. For example (repetition in bold): \textit{``Blueprint: 31\%. What percentage of women with high education are looking for an infant scale? \textbar $ $ Verbalisation (Swahili): Miongoni mwa wanawake wadogo, 31\% pekee ya wanawake walio na kipimo \textbf{cha juu cha juu cha juu cha juu...}''}

Excessive repetition in the blueprint sometimes results in the the actual verbalisation never being generated. This means the blueprint cannot be cut before comparing the output to the reference, which is highly undesirable as the blueprint itself is not meant to be evaluated against the reference.

Therefore, a repetition penalty is applied during inference (not during finetuning) to alleviate this. The repetition penalty is a form of penalised sampling, as formulated in \citet{ctrl} and simply penalises tokens which have already been generated. See \autoref{sec:appendix-rep-pen} for details on how the penalty $\theta$ is chosen.

\section{Automatic Evaluation}

Human evaluators would produce the best judgements on model outputs for T\textsc{a}TA. However, this is infeasible due to time, logistical and expense constraints. Evaluation is therefore done entirely through a combination of generic and learned automatic NLP metrics.

\subsection{\textsc{chr}F \& BLEU} \label{sec:bleuandchrf}

\textsc{chr}F and BLEU are calculated and reported for the sake of interest. However, when interpreting them it must be kept in mind that they achieve extremely low correlations with human evaluations on T\textsc{a}TA ($<$0.16) \citep{tata}.

Standard automatic evaluation metrics are so unsuitable for this task partly becasue of the reasoning required, often over multiple cells, which simply cannot be captured by these metrics, and also due to the massive verbalisation space, i.e., the number of ``correct'' verbalisations is utterly vast (especially for large tables) compared to, for example, the number of correct translations of some sentence from one language to another. This is why metrics which consider the table itself, such as PARENT \citep{parent} perform better on table-to-Text.

\subsection{F\textsc{act}KB} \label{sec:factkb}

F\textsc{act}KB differs fundamentally from metrics such as \textsc{chr}F and BLEU as it was trained specifically to evaluate whether outputs are factually faithful to their sources.

F\textsc{act}KB is intended to be used with \textit{(summarisation, article)} input pairs. We explore whether, and to what extent, it can be used to produce scores for \textit{(verbalisation, linearised table)} pairs. F\textsc{act}KB uses RoBERTa\textsubscript{Base} \citep{liu2019roberta}, which is not multilingual. Therefore it can only be used to evaluate the models finetuned on English examples.

To determine whether F\textsc{act}KB can be used to evaluate Table-to-Text on T\textsc{a}TA, scores are computed on examples from the human evaluations dataset, and the Pearson correlation between the score and the human judgement of attributability (0 or 1) is calculated. F\textsc{act}KB achieves a Pearson correlation with human evaluations of \textbf{0.22}. This outperforms the next-best existing metric, \textsc{chr}F (0.16), but is still quite low.

BERTScore \citep{zhang2020bertscore} is also considered. It only achieves a Pearson correlation of \textbf{0.12} with human evaluations on T\textsc{a}TA, which is lower than \textsc{chr}F, so it is excluded.

\subsection{S\textsc{t}ATA} \label{sec:stata}

Following instructions from \citet{tata}, a learned metric is trained on human annotations of references and model outputs for T\textsc{a}TA. It is called S\textsc{t}ATA, or \textit{Statistical Assessment of Table-to-Text in African languages}.

The human annotations dataset marks outputs as ``understandable'' if they are fluent and grammatically correct, and ``attributable'' if they satisfy the ``understandable'' criteria, and also only contain data which correctly reflects the data in at least one cell in the input table.

% \begin{wrapfigure}{R}{0.5\textwidth}
%   \centering
%   \includegraphics[width=0.5\textwidth]{ug/images/stata-corr.png}
%   \caption{Pearson correlations of various metrics with U+A, as judged by human evaluators \citep{tata}.}
%   \label{fig:stata-cor}
% \end{wrapfigure}

\citet{tata} showed that S\textsc{t}ATA performs significantly better than existing metrics for evaluating T\textsc{a}TA. \citet{tata} train three variants of the metric, but designate S\textsc{t}ATA\textsubscript{QE} (a quality estimation model which considers the linearised input and the candidate, but not the references, similar to PARENT) as the dataset's main metric as it has the highest Pearson correlation, 0.66, with the human assessments. Due to the computational constraints mentioned in \autoref{sec:appendix-train}, we use mT5\textsubscript{Large} instead of mT5\textsubscript{XXL}. This yields a Pearson correlation with human assessments of attributability of \textbf{0.59}, significant at the p $<$ 0.01 confidence level. While not as good as if the metric had been trained with mT5\textsubscript{XXL}, this is still far better than any existing metric. Since S\textsc{t}ATA\textsubscript{QE} was shown to perform the best, it is the only S\textsc{t}ATA variant we replicate.

It should be noted that the S\textsc{t}ATA scores reported cannot be directly compared to those from \citet{tata}, as they are trained with different mT5 checkpoints. They should instead be interpreted in relation to one another, and \textsc{chr}F can be used for comparison with \citet{tata} if desired.

\section{Results}
\label{sec:results}

\subsection{English Subset Results}

\begin{table*}[t]
\centering
\begin{tabular}{lllll}
\hline
\textbf{Model} & \textbf{\textsc{chr}F} & \textbf{BLEU} & \textbf{F\textsc{act}KB} & \textbf{S\textsc{t}ATA} \\ \hline
Small & 0.33 & 0.15 & 0.24 & 0.513 \\
Small (Blueprints) & 0.30 & 0.11 & 0.28 & 0.523 \\ \hline
\end{tabular}
\caption{Results for mT5\textsubscript{Small} finetuned and evaluated on the English subset of T\textsc{a}TA.}
\label{tab:eng-results}
\end{table*}

In \autoref{tab:eng-results}, \textsc{chr}F and BLEU are calculated using the predicted and reference verbalisations (traditional reference-based metrics), and F\textsc{act}KB and S\textsc{t}ATA are calculated using the predicted verbalisation and the linearised input table (Quality Estimation, or reference-less, metrics).

The results show that the blueprints do improve the attributability of outputs for the model finetuned on English examples. For the Small model, S\textsc{t}ATA increases from 0.513 to 0.523. This is significant because the range of values outputted by S\textsc{t}ATA is relatively narrow; from about 4.9 to 7.2 based on observation. This narrow range is common for learned metrics, and means that an increase of 0.1 is quite large. If the metric is trained with a larger model, the range of observed values will increase as the model becomes more confident.\footnote{mT5\textsubscript{Large} is also finetuned on the English subset, however it overfits very quickly and does not achieve representative results, due to the dataset's tiny size. The full T\textsc{a}TA dataset is small to begin with, and the English subset (about eight times smaller than the whole dataset, at 902 training examples) appears to simply be too small to finetune the Large model on. Therefore, these results are not reported.} F\textsc{act}KB also increases by 0.04 for the blueprint model.

\subsubsection{Examples \& common patterns}

Some frequent idiosyncrasies emerge upon manually examining model outputs. Models will commonly produce phrases like \textit{``increased from 15 percent to 15 percent''}. This is a common verbalisation pattern in the training data which the model learns, but it is clear that it has not learnt what \textit{``increase''} means in this context, as the numbers are the same. Similar mistakes in characterising a comparison between two numbers occur relatively often, even if both numbers do appear in the input table. For example, \textit{``mortality rate is 19, compared to 19''} This is indicative of a lack of reasoning. Note that the repetition penalty, which is applied, does not stop this.

The majority of verbalisations begin with \textit{``The percentage of...''}, \textit{``The proportion...''} or a few other common phrases. This is just a reflection of common sentence structures in the training data and is not an issue per se.

Verbalisations are also sometimes not fully formed. For example, \textit{``The proportion of children under age 5 who are wasted or too short for their age.''} or \textit{``Although the proportion of women who have received antenatal care from a skilled provider.''} These would be valid as the first half of verbalisations, but are not completed and therefore do not make sense.

Generally, the title and unit parts of the input tables seem to appear the most consistently and accurately in the verbalisations, with the actual data points less so. This is consistent with an observation made by \citet{tata}.

\subsection{Multilingual Results}

\begin{table}[h]
\begin{tabular}{llll}
\hline
\textbf{Model} & \textbf{\textsc{chr}F} & \textbf{BLEU} & \textbf{S\textsc{t}ATA} \\ \hline
Small & 0.32 & 0.16 & 0.552 \\
Small (Eng blu.) & 0.29 & 0.09 & 0.525 \\
Small (Trans blu.) & 0.30 & 0.12 & 0.542 \\
Large & 0.33 & 0.13 & 0.552 \\
Large (Eng blu.) & 0.24 & 0.04 & 0.519 \\
Large (Trans blu.) & 0.27 & 0.11 & 0.544 \\ \hline
\end{tabular}
\caption{Results of finetuned multilingual mT5 models on the test set (all languages). Models with "blu." are models with blueprints.}
\label{tab:results}
\end{table}

The Small baseline model trained by \citet{tata} achieves a \textsc{chr}F of 0.33, which is very close to the 0.32 achieved by our baseline Small model (\autoref{tab:results}), so the results are considered to be replicated.

The English blueprint setup performs poorly (\autoref{tab:results}). Inspecting outputs of models trained on the English blueprints, which have to be able to generate outputs that contain multiple languages (English and the target language) reveals that they sometimes mix languages up. In the following example, colours denote different languages: \textit{\color{ForestGreen} ``Among Tanzania, one-third of Tanzanian women in the United States had all the three or more antenatal care visits, \color{Plum} asilimia 29 ya wanawake \color{Green} in the United States had all the three or more antenatal care visits.''}

\color{black}
\textit{``asilimia 29 ya wanawake''} is a Swahili phrase meaning \textit{``29 percent of women''} which has been incorrectly generated in the middle of an English sentence. (In this example, the United States is also hallucinated and a clause is repeated, but those are separate issues).

The translated blueprints fare decisively better than the English ones across all metrics, but are still slightly worse than no blueprints (\autoref{tab:results}).

\subsubsection{Model size}

\citet{tata} saw a large performance jump on T\textsc{a}TA from mT5\textsubscript{Small} to mT5\textsubscript{XXL} (13B). However, we do not observe Large (1.2B) improving over Small. One likely hypothesis is model-wise double descent, the phenomenon whereby performance degrades as the model size increases to a point, then begins to improve again as the model size is increased further \citep{doubledescent}. Even the full T\textsc{a}TA dataset is small, and the Large model converges quickly or overfits. Increasing the model size by a factor of 10 would still result in a fast convergence, but the model will likely achieve a much lower loss before this happens. We are unable to test this hypothesis due to compute restraints.

\subsection{Per-Language Analysis}

\begin{table*}[t]
\centering
\begin{tabular}{lcccc}
\hline
\textbf{Lang} & \textbf{Small} & \textbf{Small Blueprints} & \textbf{Large} & \textbf{Large Blueprints} \\ \hline
% \multirow{2}{*}{\textbf{Language}} & \multicolumn{2}{c}{\textbf{Small}} & \multicolumn{2}{c}{\textbf{Large}} \\
%  & \textbf{Vanilla} & \textbf{Blueprint} & \textbf{Vanilla} & \textbf{Blueprint} \\ \hline
En & \textbf{0.19} / 0.33 / \textbf{0.551} & 0.15 / 0.34 / 0.529 & 0.17 / \textbf{0.37} / 0.538 & 0.10 / 0.29 / 0.549 \\
Sw & \textbf{0.21} / \textbf{0.39} / \textbf{0.589} & 0.17 / 0.36 / 0.569 & 0.16 / 0.38 / 0.581 & 0.13 / 0.30 / 0.585 \\
Yo & \textit{\textbf{0.03}} / 0.13 / 0.567 & \textit{\textbf{0.03}} / \textit{\textbf{0.14}} / 0.563 & \textit{\textbf{0.03}} / \textit{\textbf{0.14}} / \textbf{0.577} & 0.02 / \textit{\textbf{0.14}} / 0.561 \\
Fr & \textbf{0.17} / 0.36 / 0.528 & 0.11 / 0.33 / 0.526 & 0.14 / \textbf{0.38} / 0.526 & 0.12 / 0.31 / \textbf{0.529} \\
Pt & \textbf{0.17} / \textit{\textbf{0.39}} / 0.527 & 0.16 / 0.34 / 0.518 & 0.15 / \textit{\textbf{0.39}} / 0.512 & 0.15 / 0.32 / \textbf{0.531} \\
Ha & \textbf{0.17} / \textit{\textbf{0.33}} / 0.526 & 0.12 / \textit{\textbf{0.33}} / 0.523 & 0.12 / \textit{\textbf{0.33}} / \textbf{0.546} & 0.12 / 0.29 / 0.515 \\
Ar & \textbf{0.14} / 0.32 / \textbf{0.539} & 0.11 / \textit{\textbf{0.33}} / 0.523 & 0.12 / \textit{\textbf{0.33}} / 0.533 & 0.12 / 0.31 / 0.519 \\
Ig & \textbf{0.20} / \textbf{0.35} / 0.596 & 0.17 / 0.32 / 0.584 & 0.16 / 0.34 / \textbf{0.605} & 0.15 / 0.27 / 0.558 \\ \hline
\end{tabular}
\caption{Language-specific performance of mT5 multilingual models (BLEU / \textsc{chr}F / S\textsc{t}ATA). Bold figures represent the best result for each metric in each row. Italic figures are ties.}
\label{tab:lang-eval}
\end{table*}

For more granular insights, the multilingual models are evaluated on each language in the test set individually. (The Blueprint columns in this table are translated blueprints as these were shown to perform better in \autoref{tab:results}). These per-language evaluation results are noisy (\autoref{tab:lang-eval}). Broadly, blueprints rarely improve any of the metrics, and performance between the Small and Large model is very close.

The lower-resource African languages also tend to benefit the most from an increase in model size. Igbo, Hausa and Yor\`ub\'a all achieve higher S\textsc{t}ATA scores with the Large model, and are the only languages to do so. This suggests that scale is of particular importance for the low-resource languages.

\subsection{Blueprint Analysis}

\begin{table*}[t]
\centering
\begin{tabular}{lll}
\hline
\textbf{Model} & \textbf{\textsc{chr}F} & \textbf{BLEU} \\ \hline
Small Trans Blueprints (Multilingual) & 0.27 & 0.07 \\
Small Blueprints (English) & 0.23 & 0.05 \\ \hline
\end{tabular}
\caption{\textsc{chr}F and BLEU between predicted and gold blueprints on dev set.}
\label{tab:blueprint-predicted}
\end{table*}

\autoref{tab:blueprint-predicted} shows how closely predicted blueprints match reference blueprints in the dev set (the candidates and references are split on \textit{``Verbalisation:''} and only the blueprints compared). The dev set is used as no blueprints are generated for the test set, because at test time only the verbalisations are compared.

These scores are very low. It should be noted that achieving a high \textsc{chr}F or BLEU on the blueprints, or indeed the entire output, is not the explicit goal of training. There are many valid blueprints for any given table, especially when generating short verbalisations from large tables, where there is lots of data to choose from. Arguably more important, is whether blueprints are related to the input table, and whether verbalisations are related to their blueprints.

\autoref{tab:blueprint-similarity} provides a measure of this. It shows how well the models are able to generate blueprints, and to what extent the verbalisations use these blueprints. To do this, \textsc{chr}F and BLEU are calculated between the linearised input and the blueprint to quantify how much information from the table is present in the blueprint. \textsc{chr}F and BLEU are also calculated between the blueprint and the verbalisation to quantify how closely the output relies on the blueprint for content selection.

Again, note that the goal is obviously not to have blueprints which are exactly the same as the input tables, nor verbalisations which are exactly the same as the blueprints. So, just maximising these metrics is not desirable here. The models' scores are to be interpreted with respect to the dataset scores, as an indicator the extent to which the models exhibit desirable attributes of the dataset.

\begin{table*}[t]
\centering
\begin{tabular}{lllll}
\hline
 & \multicolumn{2}{c}{\textbf{Linearised input $\rightarrow$ Blueprint}} & \multicolumn{2}{c}{\textbf{Blueprint $\rightarrow$ Verbalisation}} \\
 & \makebox[1.5cm][c]{\textbf{\textsc{chr}F}\hfill} & \textbf{BLEU} & \makebox[1.5cm][c]{\textbf{\textsc{chr}F}\hfill} & \textbf{BLEU} \\ \hline
English dataset & 0.28 & 0.02 & 0.61 & 0.24 \\
English model & 0.24 & 0.02 & 0.39 & 0.20 \\
Multilingual dataset & 0.25 & 0.02 & 0.43 & 0.13 \\
Multilingual model & 0.23 & 0.01 & 0.36 & 0.16 \\ \hline
\end{tabular}
\caption{\textsc{chr}F and BLEU between linearised inputs, blueprints and verbalisations in the training data and model outputs.}
\label{tab:blueprint-similarity}
\end{table*}

The \textbf{English dataset} and \textbf{Multilingual dataset} rows in \autoref{tab:blueprint-similarity} represent scores calculated directly on the respective training datasets, and showcase the best-case scores. The \textbf{English model} (mT5\textsubscript{Small} finetuned on the English subset with blueprints) and \textbf{Multilingual model} (mT5\textsubscript{Small} finetuned on the full dataset with translated blueprints) rows represent scores calculated on the respective test sets, and the models' generated blueprints and verbalisations (the generation is split on \textit{``Verbalisation:''} and the blueprint and verbalisation separated).

Even the English model is not able to generate blueprints that are as closely related to the linearised table as the references (\textsc{chr}F of 0.24 versus 0.28). Furthermore, the similarity between the blueprints and the outputs is very low (\textsc{chr}F 0.39, BLEU 0.20) compared to the dataset (\textsc{chr}F 0.61, BLEU 0.24). This indicates that not only is the model struggling to generate blueprints that are as good as gold, but it is also failing to remain as faithful to its blueprint (as showcased in the second example in \ref{tab:output-eg}). In other words, the model's verbalisations have significantly less in common with its blueprints than the dataset's.

The multilingual model actually does a better job of producing verbalisations which rely on the blueprints. There is a smaller percentage drop in \textbf{Blueprint $\rightarrow$ Verbalisation} \textsc{chr}F than for English, and BLEU actually improves slightly. It is clear from these findings that the model's verbalisations do not draw from its blueprints enough, as exemplified in the examples in \autoref{tab:output-eg}. One way of encouraging the model to rely more on its blueprints would be to use a form of constrained decoding, which could help focus the model on using words which it generates in its blueprint.\footnote{\url{https://huggingface.co/blog/constrained-beam-search}} This is not explored and is left to future work.

Still, there is a fundamental disadvantage in the multilingual setup before training even begins. Note how much lower the multilingual dataset's \textbf{Blueprint $\rightarrow$ Verbalisation} score (\textsc{chr}F 0.43, BLEU 0.13) is than the English dataset's (\textsc{chr}F 0.61, BLEU 0.24). This indicates that, as predicted in \autoref{sec:making-mult}, inaccuracies in translating the English blueprints into the target languages have made it challenging to create high-quality multilingual blueprints.

So, why do the multilingual blueprint models achieve slightly higher S\textsc{t}ATA scores than the English-only blueprint models if the multilingual blueprints are lower quality? These multilingual models are still trained on around eight times more training data, as they see the full dataset, not just the English subset. This allows the multilingual models to learn the general task better, despite the language-specific challenges, and is also the reason the multilingual model achieves higher BLEU and \textsc{chr}F between predicted and reference blueprints in \autoref{tab:blueprint-predicted}.

\section{Conclusions}

The evidence suggests QA blueprints are effectively improve the attributability of Table-to-Text outputs in English. More work needs to be conducted to validate this, with both a larger English Table-to-Text dataset such as ToTTo, and a larger model, such as mT5\textsubscript{XXL} or the newly-released Gemma-7b.\footnote{\url{https://blog.google/technology/developers/gemma-open-models/}}

In the multilingual setup, English blueprints degrade performance significantly and sometimes cause models to mix up multiple languages in their verbalisations.

Translated blueprints fare better, but still worse than no blueprints at all. This is due to inevitable inaccuracies in machine translating the blueprints from English to the target languages in the dataset for training. These imperfect translated blueprints mean that the multilingual models do not have high-quality gold examples to learn from.

As a result, the models struggle to generate blueprints which are as closely related to the input tables as the dataset. This problem, as measured by BLEU, is more severe for multilingual models than English.

For the English results, an increase in S\textsc{t}ATA scores is observed, while generic automated metrics such as BLEU and \textsc{chr}F decrease. This further confirms that these metrics are not suitable for evaluating T\textsc{a}TA due to their very low correlations with human evaluations. Although F\textsc{act}KB performs slightly better, is not recommended for evaluating T\textsc{a}TA in future work, as its correlation with humans is still quite low, and this is not the task it was designed for. The learned metric S\textsc{t}ATA should be the definitive judge of model performance on T\textsc{a}TA when human evaluators are not available. We release our version of S\textsc{t}ATA online in the hope that it will make doing research on T\textsc{a}TA more accessible. However, this version is trained with mT5\textsubscript{Large}. It should ideally be retrained with mT5\textsubscript{XXL}, released and standardised, else comparing scores across papers will be impossible.

It is also observed that increasing model size results in larger gains for the low-resource languages.

Multilingual Table-to-Text generation remains a very challenging task for neural models. Based on our findings, recommendations for future work are as follows: Trial using LLMs to generate more synthetic training data in several languages. If high-quality synthetic data can be generated, and the dataset size increased, this will boost model performance (although LLMs will probably only be able to generate good examples in the higher-resource languages). Additionally, constrained decoding could also be explored as a way to make verbalisations utilise the blueprints more. Finally, T\textsc{a}TA can be turned into a more constrained task by having human annotators highlight the cells in each table that are used in its reference verbalisation, as the ToTTo dataset does. This greatly reduces the valid answer space, especially for larger tables.

\section{Acknowledgements}

This paper is a condensed version of my undergraduate dissertation. I would like to thank my outstanding supervisor, Prof. Mirella Lapata, for her guidance and support. To be supervised by such a distinguished researcher was a privilege.

\bibliography{custom}

\begin{thebibliography}{21}
\providecommand{\natexlab}[1]{#1}

\bibitem[{Chen et~al.(2023)Chen, Wang, Chen, Yu, Ma, Zhao, Zhang, and Yu}]{chen2023dense}
Tong Chen, Hongwei Wang, Sihao Chen, Wenhao Yu, Kaixin Ma, Xinran Zhao, Hongming Zhang, and Dong Yu. 2023.
\newblock \href {https://arxiv.org/abs/2312.06648} {Dense x retrieval: What retrieval granularity should we use?}
\newblock \emph{Preprint}, arXiv:2312.06648.

\bibitem[{Chung et~al.(2022)Chung, Hou, Longpre, Zoph, Tay, Fedus, Li, Wang, Dehghani, Brahma, Webson, Gu, Dai, Suzgun, Chen, Chowdhery, Castro-Ros, Pellat, Robinson, Valter, Narang, Mishra, Yu, Zhao, Huang, Dai, Yu, Petrov, Chi, Dean, Devlin, Roberts, Zhou, Le, and Wei}]{chung2022scaling}
Hyung~Won Chung, Le~Hou, Shayne Longpre, Barret Zoph, Yi~Tay, William Fedus, Yunxuan Li, Xuezhi Wang, Mostafa Dehghani, Siddhartha Brahma, Albert Webson, Shixiang~Shane Gu, Zhuyun Dai, Mirac Suzgun, Xinyun Chen, Aakanksha Chowdhery, Alex Castro-Ros, Marie Pellat, Kevin Robinson, Dasha Valter, Sharan Narang, Gaurav Mishra, Adams Yu, Vincent Zhao, Yanping Huang, Andrew Dai, Hongkun Yu, Slav Petrov, Ed~H. Chi, Jeff Dean, Jacob Devlin, Adam Roberts, Denny Zhou, Quoc~V. Le, and Jason Wei. 2022.
\newblock \href {https://arxiv.org/abs/2210.11416} {Scaling instruction-finetuned language models}.
\newblock \emph{Preprint}, arXiv:2210.11416.

\bibitem[{Dhingra et~al.(2019)Dhingra, Faruqui, Parikh, Chang, Das, and Cohen}]{parent}
Bhuwan Dhingra, Manaal Faruqui, Ankur Parikh, Ming-Wei Chang, Dipanjan Das, and William~W. Cohen. 2019.
\newblock \href {https://arxiv.org/abs/1906.01081} {Handling divergent reference texts when evaluating table-to-text generation}.
\newblock \emph{Preprint}, arXiv:1906.01081.

\bibitem[{Feng et~al.(2023)Feng, Balachandran, Bai, and Tsvetkov}]{feng2023factkb}
Shangbin Feng, Vidhisha Balachandran, Yuyang Bai, and Yulia Tsvetkov. 2023.
\newblock \href {https://arxiv.org/abs/2305.08281} {Factkb: Generalizable factuality evaluation using language models enhanced with factual knowledge}.
\newblock \emph{Preprint}, arXiv:2305.08281.

\bibitem[{Gehrmann et~al.(2022)Gehrmann, Ruder, Nikolaev, Botha, Chavinda, Parikh, and Rivera}]{tata}
Sebastian Gehrmann, Sebastian Ruder, Vitaly Nikolaev, Jan~A. Botha, Michael Chavinda, Ankur Parikh, and Clara Rivera. 2022.
\newblock \href {https://arxiv.org/abs/2211.00142} {Tata: A multilingual table-to-text dataset for african languages}.
\newblock \emph{Preprint}, arXiv:2211.00142.

\bibitem[{Kale and Rastogi(2020)}]{kale-rastogi-2020-text}
Mihir Kale and Abhinav Rastogi. 2020.
\newblock \href {https://doi.org/10.18653/v1/2020.inlg-1.14} {Text-to-text pre-training for data-to-text tasks}.
\newblock In \emph{Proceedings of the 13th International Conference on Natural Language Generation}, pages 97--102, Dublin, Ireland. Association for Computational Linguistics.

\bibitem[{Keskar et~al.(2019)Keskar, McCann, Varshney, Xiong, and Socher}]{ctrl}
Nitish~Shirish Keskar, Bryan McCann, Lav~R. Varshney, Caiming Xiong, and Richard Socher. 2019.
\newblock \href {https://arxiv.org/abs/1909.05858} {{CTRL:} {A} conditional transformer language model for controllable generation}.
\newblock \emph{CoRR}, abs/1909.05858.

\bibitem[{Liu et~al.(2019)Liu, Ott, Goyal, Du, Joshi, Chen, Levy, Lewis, Zettlemoyer, and Stoyanov}]{liu2019roberta}
Yinhan Liu, Myle Ott, Naman Goyal, Jingfei Du, Mandar Joshi, Danqi Chen, Omer Levy, Mike Lewis, Luke Zettlemoyer, and Veselin Stoyanov. 2019.
\newblock \href {https://arxiv.org/abs/1907.11692} {Roberta: A robustly optimized bert pretraining approach}.
\newblock \emph{Preprint}, arXiv:1907.11692.

\bibitem[{Manakul et~al.(2023)Manakul, Liusie, and Gales}]{mqag}
Potsawee Manakul, Adian Liusie, and Mark J.~F. Gales. 2023.
\newblock \href {https://arxiv.org/abs/2301.12307} {Mqag: Multiple-choice question answering and generation for assessing information consistency in summarization}.
\newblock \emph{Preprint}, arXiv:2301.12307.

\bibitem[{Nakkiran et~al.(2019)Nakkiran, Kaplun, Bansal, Yang, Barak, and Sutskever}]{doubledescent}
Preetum Nakkiran, Gal Kaplun, Yamini Bansal, Tristan Yang, Boaz Barak, and Ilya Sutskever. 2019.
\newblock \href {https://arxiv.org/abs/1912.02292} {Deep double descent: Where bigger models and more data hurt}.
\newblock \emph{CoRR}, abs/1912.02292.

\bibitem[{Narayan et~al.(2023)Narayan, Maynez, Amplayo, Ganchev, Louis, Huot, Sandholm, Das, and Lapata}]{qa-generation}
Shashi Narayan, Joshua Maynez, Reinald~Kim Amplayo, Kuzman Ganchev, Annie Louis, Fantine Huot, Anders Sandholm, Dipanjan Das, and Mirella Lapata. 2023.
\newblock \href {https://arxiv.org/abs/2207.00397} {Conditional generation with a question-answering blueprint}.
\newblock \emph{Preprint}, arXiv:2207.00397.

\bibitem[{Papineni et~al.(2002)Papineni, Roukos, Ward, and Zhu}]{bleu}
Kishore Papineni, Salim Roukos, Todd Ward, and Wei-Jing Zhu. 2002.
\newblock \href {https://doi.org/10.3115/1073083.1073135} {{B}leu: a method for automatic evaluation of machine translation}.
\newblock In \emph{Proceedings of the 40th Annual Meeting of the Association for Computational Linguistics}, pages 311--318, Philadelphia, Pennsylvania, USA. Association for Computational Linguistics.

\bibitem[{Parikh et~al.(2020)Parikh, Wang, Gehrmann, Faruqui, Dhingra, Yang, and Das}]{totto}
Ankur Parikh, Xuezhi Wang, Sebastian Gehrmann, Manaal Faruqui, Bhuwan Dhingra, Diyi Yang, and Dipanjan Das. 2020.
\newblock \href {https://doi.org/10.18653/v1/2020.emnlp-main.89} {{ToTTo}: A controlled table-to-text generation dataset}.
\newblock In \emph{Proceedings of the 2020 Conference on Empirical Methods in Natural Language Processing (EMNLP)}, pages 1173--1186, Online. Association for Computational Linguistics.

\bibitem[{Popovi{\'c}(2015)}]{chrf}
Maja Popovi{\'c}. 2015.
\newblock \href {https://doi.org/10.18653/v1/W15-3049} {chr{F}: character n-gram {F}-score for automatic {MT} evaluation}.
\newblock In \emph{Proceedings of the Tenth Workshop on Statistical Machine Translation}, pages 392--395, Lisbon, Portugal. Association for Computational Linguistics.

\bibitem[{Puduppully et~al.(2019)Puduppully, Dong, and Lapata}]{content-planning}
Ratish Puduppully, Li~Dong, and Mirella Lapata. 2019.
\newblock \href {https://doi.org/10.1609/aaai.v33i01.33016908} {Data-to-text generation with content selection and planning}.
\newblock In \emph{Proceedings of the Thirty-Third AAAI Conference on Artificial Intelligence and Thirty-First Innovative Applications of Artificial Intelligence Conference and Ninth AAAI Symposium on Educational Advances in Artificial Intelligence}, AAAI'19/IAAI'19/EAAI'19. AAAI Press.

\bibitem[{Rajpurkar et~al.(2016)Rajpurkar, Zhang, Lopyrev, and Liang}]{rajpurkar-etal-2016-squad}
Pranav Rajpurkar, Jian Zhang, Konstantin Lopyrev, and Percy Liang. 2016.
\newblock \href {https://doi.org/10.18653/v1/D16-1264} {{SQ}u{AD}: 100,000+ questions for machine comprehension of text}.
\newblock In \emph{Proceedings of the 2016 Conference on Empirical Methods in Natural Language Processing}, pages 2383--2392, Austin, Texas. Association for Computational Linguistics.

\bibitem[{Roberts et~al.(2019)Roberts, Raffel, Lee, Matena, Shazeer, Liu, Narang, Li, and Zhou}]{t5}
Adam Roberts, Colin Raffel, Katherine Lee, Michael Matena, Noam Shazeer, Peter~J. Liu, Sharan Narang, Wei Li, and Yanqi Zhou. 2019.
\newblock Exploring the limits of transfer learning with a unified text-to-text transformer.
\newblock Technical report, Google.

\bibitem[{Ruder(2022)}]{ruder2022statemultilingualai}
Sebastian Ruder. 2022.
\newblock {The State of Multilingual AI}.
\newblock \url{http://ruder.io/state-of-multilingual-ai/}.

\bibitem[{Wiseman et~al.(2017)Wiseman, Shieber, and Rush}]{wiseman-etal-2017-challenges}
Sam Wiseman, Stuart Shieber, and Alexander Rush. 2017.
\newblock \href {https://doi.org/10.18653/v1/D17-1239} {Challenges in data-to-document generation}.
\newblock In \emph{Proceedings of the 2017 Conference on Empirical Methods in Natural Language Processing}, pages 2253--2263, Copenhagen, Denmark. Association for Computational Linguistics.

\bibitem[{Xue et~al.(2021)Xue, Constant, Roberts, Kale, Al-Rfou, Siddhant, Barua, and Raffel}]{mt5}
Linting Xue, Noah Constant, Adam Roberts, Mihir Kale, Rami Al-Rfou, Aditya Siddhant, Aditya Barua, and Colin Raffel. 2021.
\newblock \href {https://doi.org/10.18653/v1/2021.naacl-main.41} {m{T}5: A massively multilingual pre-trained text-to-text transformer}.
\newblock In \emph{Proceedings of the 2021 Conference of the North American Chapter of the Association for Computational Linguistics: Human Language Technologies}, pages 483--498, Online. Association for Computational Linguistics.

\bibitem[{Zhang et~al.(2020)Zhang, Kishore, Wu, Weinberger, and Artzi}]{zhang2020bertscore}
Tianyi Zhang, Varsha Kishore, Felix Wu, Kilian~Q. Weinberger, and Yoav Artzi. 2020.
\newblock \href {https://arxiv.org/abs/1904.09675} {Bertscore: Evaluating text generation with bert}.
\newblock \emph{Preprint}, arXiv:1904.09675.

\end{thebibliography}

\appendix

\section{Model Training Details}
\label{sec:appendix-train}

\citet{tata} use mT5\textsubscript{Small} and mT5\textsubscript{XXL}. However, mT5\textsubscript{XXL}, and indeed mT5\textsubscript{XL} are both massive models (13B and 3.7B parameters respectively). Available compute resources made finetuning these models infeasible. The largest model that could be trained is mT5\textsubscript{Large} (1.2B), so this is used instead.

Both mT5\textsubscript{Small} and mT5\textsubscript{Large} are finetuned with the following hyperparameters and setup: a constant learning rate of 0.001, dropout of 0.1, per-device batch size of 4, 5 epochs for the Small model and 3 for Large, weight decay of 0.001 for Large, linearised tables (inputs) and references are both truncated to 512 tokens, and validation loss is measured every 100 or 500 steps depending on batch size and after training the checkpoint with the lowest loss is selected.

\section{S\textsc{t}ATA Training Details}
\label{sec:appendix-stata}

mT5\textsubscript{Large} and mT5\textsubscript{Small} are finetuned for regression with an RMSE (Root Mean Squared Error) loss function. The metric is released on the Huggingface Hub\footnote{\url{https://huggingface.co/adenhaus/mt5-large-stata}} in the hope that it will prove useful to other researchers who wish to work on T\textsc{a}TA.

A spare token in mT5's vocabulary is chosen as the regression token. At each step, the RMSE of this token's logit and the label, which is 0 or 1, is taken. This is the loss.\footnote{\url{https://medium.com/p/a71d12aaf075}}

\begin{equation}
    RMSE(y, \hat{y}) = \sqrt{\frac{\sum_{i=0}^{N - 1} (y_i - \hat{y}_i)^2}{N}}
    \label{eq:rmse}
\end{equation}

In \autoref{eq:rmse}, $\hat{y}$ is the logit of the special token and $y$ is the label. $N$ is the number of observations. At inference time, generation is constrained to this token, and its logit $x$ is converted into a probability with the sigmoid, or logistic function (\autoref{eq:logistic}). This is the final metric.

\begin{equation}
    \sigma (x) =  \frac{\mathrm{1} }{\mathrm{1} + e^{-x} }
    \label{eq:logistic}
\end{equation}

The human annotations file contains rows with the following fields and data types: output (string), model (string), interpretable (float), attributable (float), cells (float), reasoning (float), id (string), set (string), lang (string) and linearized\_input (string).

\begin{figure}[h]
  \centering
  \includegraphics[width=0.8\columnwidth]{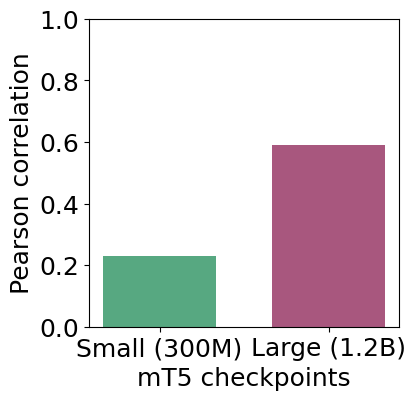}
  \caption{Pearson correlations of mT5-Small and mT5-Large S\textsc{t}ATA models with U+A human evaluations.}
  \label{fig:stata-checkpoint-corr}
\end{figure}

The file required some cleaning before it could be used for training. All rows where ``attributable'' is not either 0.0 or 1.0 are dropped. 73.7\% of samples have a 0 in the ``attributable'' column, and 26.3\% have a 1. This phenomenon, where the dataset is unbalanced in favour of the negative class, is common in classification or similar datasets. However, since the model learns the data quite effectively as-is, no data augmentation or other techniques are applied to balance the classes. The file is then randomly sampled into train (80\%), validation (10\%) and test (10\%) splits. This makes the training set 4,900 rows and the validation and test sets 612 rows each. This dataset is also released on the Hugging Face Hub\footnote{\url{https://huggingface.co/datasets/adenhaus/stata}} so that S\textsc{t}ATA can be easily retrained by other researchers, particularly those with more compute resources who can train it with mT5\textsubscript{XXL}.

To evaluate the metric, scores are computed on the test set and the Pearson correlation between them and the human assessments of attributability (1, or 0) is computed. \autoref{fig:stata-checkpoint-corr} shows the correlations for the Small and Large model.

\subsection{Hyperparameters \& other details}

Models are finetuned with the following setup: a constant learning rate of 0.0001, per-device batch size of 32 for Small and 8 for Large, 15 epochs, dropout of 0.1, inputs are truncated to 2048 tokens, validation loss is measured every 0.1 epochs, and after training the checkpoint with the lowest RMSE loss is selected.

\section{Repetition Penalty Details}
\label{sec:appendix-rep-pen}

This is simply Hugging Face's \textit{repetition\_penalty}\footnote{\url{https://huggingface.co/docs/transformers/internal/generation_utils}} parameter which can be passed to \textit{model.generate}.

\begin{figure}[h]
  \includegraphics[width=\columnwidth]{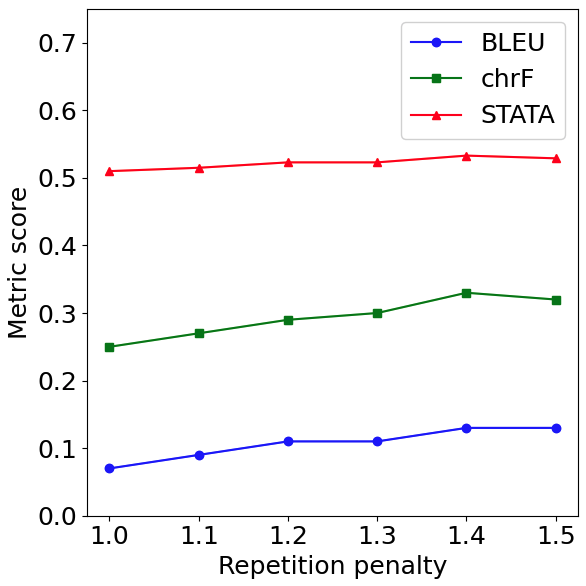}
  \caption{Multilingual model performance on the test set (including all languages) with varying repetition penalties.}
  \label{fig:rep-pen}
\end{figure}

An analysis of varying repetition penalties is performed on the English-finetuned mT5\textsubscript{Small} with blueprints, as the model and dataset are small which allows for a faster analysis. \autoref{fig:rep-pen} shows that as the repetition penalty is increased, performance on all metrics increases until 1.4, after which the metrics begin to decline. A $\theta$ of 1.0 is equivalent to no penalty. However, by inspecting outputs it is observed that a penalty of 1.2 is enough to stop most of the highly problematic repetitive outputs. The penalty is somewhat a blunt instrument, so it is not over-used here. Although the scores are higher, a penalty of 1.2 is cautiously chosen for the rest of the experiments as it mitigates the main problem, without altering the normal generations excessively. This agrees with the findings of \citet{ctrl}, who also note that a $\theta$ of 1.2 strikes a good balance between keeping repetition to a minimum but maintaining fluent and sensible outputs.

Where highly repetitive blueprints do still occur at test time, these obviously bring down the metrics slightly, as a blueprint is being compared to a reference verbalisation which is not intended. But these bad candidates are not removed.

\section{Blueprint Translation Example}
\label{sec:appendix-translations}

The blueprint, generated in English (see \autoref{sec:appendix-blu-ex} for full examples), is:

\begin{quote}
    \textit{13\%. How much of young women in Mali are in the highest tercile for empowerment? \textbar 81\%. What is the percentage of young women in the Philippines who are in the highest tercile for empowerment? \textbar}
\end{quote}

This is translated into Swahili as:

\begin{quote}
    \textit{13\%. Je, ni idadi gani ya wanawake vijana nchini Mali walio katika eneo la juu zaidi la kuwezeshwa? \textbar $ $ 81\%. Je, ni asilimia ngapi ya wanawake vijana nchini Ufilipino walio katika eneo la juu zaidi la kuwezeshwa? \textbar}
\end{quote}

Translated back into English (again, with Google Translate) to check the quality:

\begin{quote}
    \textit{13\%. What is the proportion of young women in Mali in the highest area of empowerment? \textbar $ $ 81\%. What percentage of young women in the Philippines are in the highest empowerment zone? \textbar}
\end{quote}

This is relatively close to the original English blueprint, however it loses the term ``tercile'' and replaces it with ``area'' or ``zone'', which could have many different meanings (e.g. spatial) that do not closely relate to ``tercile''. Swahili is one of the languages with better translation performance. For those with even poorer translations, the blueprints will not match perfectly to the verbalisations, which are translated by expert humans \citep{tata}.

\section{Blueprint Examples}
\label{sec:appendix-blu-ex}

See \autoref{tab:examples}.

\begin{table*}[t]
\centering
\begin{tabular}{p{0.1\linewidth}p{0.17\linewidth}p{0.22\linewidth}p{0.35\linewidth}}
\hline
\textbf{Dataset} & \textbf{Source verbalisation} & \textbf{Generated blueprint} & \textbf{New reference} \\ \hline
\color{black} English & \color{Brown} Only 13\% of young women in Mali are in the highest tercile for empowerment compared with 81\% of young women in the Philippines. & \color{Fuchsia} 13\%. How much of young women in Mali are in the highest tercile for empowerment? \textbar 81\%. What is the percentage of young women in the Philippines who are in the highest tercile for empowerment? \textbar & \color{black} Blueprint: \color{Fuchsia} 13\%. How much of young women in Mali are in the highest tercile for empowerment? \textbar 81\%. What is the percentage of young women in the Philippines who are in the highest tercile for empowerment? \textbar \color{black} Verbalisation: \color{Brown} Only 13\% of young women in Mali are in the highest tercile for empowerment compared with 81\% of young women in the Philippines. \color{black} \\ \hline
\color{black} Translated blueprints & \color{Brown} Asilimia 13 pekee ya wanawake wadogo katika Mali ndio wapo katika kikundi cha juu cha uwezeshaji ikilinganishwa na asilimia 81 ya wanawake wadogo katika Ufilipino. & \color{Fuchsia} 13\%. Je, ni idadi gani ya wanawake vijana nchini Mali walio katika eneo la juu zaidi la kuwezeshwa? \textbar 81\%. Je, ni asilimia ngapi ya wanawake vijana nchini Ufilipino walio katika eneo la juu zaidi la kuwezeshwa? \textbar & \color{black} Blueprint: \color{Fuchsia} 13\%. Je, ni idadi gani ya wanawake vijana nchini Mali walio katika eneo la juu zaidi la kuwezeshwa? \textbar 81\%. Je, ni asilimia ngapi ya wanawake vijana nchini Ufilipino walio katika eneo la juu zaidi la kuwezeshwa? \textbar \color{black} Verbalisation: \color{Brown} Asilimia 13 pekee ya wanawake wadogo katika Mali ndio wapo katika kikundi cha juu cha uwezeshaji ikilinganishwa na asilimia 81 ya wanawake wadogo katika Ufilipino. \color{black} \\ \hline
\color{black} English blueprints & \color{Brown} Seules 13\% des jeunes femmes au Mali se situent dans le tercile le plus élevé en matière d'autonomisation, contre 81\% des jeunes femmes aux Philippines. & \color{Fuchsia} 13\%. How much of young women in Mali are in the highest tercile for empowerment? \textbar 81\%. What is the percentage of young women in the Philippines who are in the highest tercile for empowerment? \textbar & \color{black} Blueprint: \color{Fuchsia} 13\%. How much of young women in Mali are in the highest tercile for empowerment? \textbar 81\%. What is the percentage of young women in the Philippines who are in the highest tercile for empowerment? \textbar \color{black} Verbalisation (French): \color{Brown} Seules 13\% des jeunes femmes au Mali se situent dans le tercile le plus élevé en matière d'autonomisation, contre 81\% des jeunes femmes aux Philippines \color{black}. \color{black} \\ \hline
\end{tabular}
\caption{Example blueprints from the three dataset setups for the same input table with English, Swahili and French as target languages.}
\label{tab:examples}
\end{table*}

\section{Output Examples}
\label{sec:appendix-out}

In the first example in \autoref{tab:output-eg}, neither the vanilla nor the blueprint verbalisations are very good in terms of understandability or attributability. Both refer to the correct concepts, but the former makes no reference to any actual data, and the second entirely hallucinates the data. The blueprint is semi-correct, as ``wanted then'' is the most common category, at 0.57. Unfortunately, this data is not correctly referenced in the verbalisation at all.

In the second example, a very good blueprint has been formulated. It captures important information from the table and does so correctly, identifying the right figure (88\%) and year (2003). However, again the verbalisation fails to use this blueprint effectively, mentioning the correct year, but not the figure. This is still an improvement over the vanilla verbalisation, which references neither.

\begin{table*}[t]
\centering
\begin{tabular}{p{0.15\linewidth} p{0.22\linewidth} p{0.1\linewidth} p{0.2\linewidth} p{0.2\linewidth}}
\hline
\textbf{Reference} & \textbf{Linearised input} & \textbf{Setup} & \textbf{Blueprint} & \textbf{Verbalisation} \\ \hline
\multirow{2}{1\linewidth}{17 percent of births in Kenya are unwanted, and 26 percent are mistimed (wanted later).} & \multirow{2}{1\linewidth}{\color{black}Planning Status of \color{Red}Births \color{black} \textbar Percent \textbar (\color{Tan}Wanted \color{black} then, 0.57) (Unwanted, 0.17) (\color{Tan}Wanted \color{black} later, 0.26)} & Vanilla &  & The proportion of \color{Red}births \color{black} were \color{Tan}wanted \color{black} at the time of delivery. \\ \cline{3-5}
 &  & \color{black}Blueprint & \color{Tan}wanted \color{Red}births\color{black}. What was the most common form of \color{Red}births\color{black}? & \color{black}The proportion of \color{Red}births \color{black} were \color{Tan}wanted \color{black} at \color{Red}births\color{black}, from a low of 7\% in 1990 to a high of 11\% in 2012. \\[0.5cm] \hline
\multirow{2}{1\linewidth}{The 2008-09 data indicate a rise since 2003 in medical antenatal care coverage.} & \multirow{2}{1\linewidth}{\color{black} Trends in Receipt of \color{JungleGreen}Antenatal Care from a Skilled Medical Provider \color{black}, Kenya \color{Tan}2003-2008 \color{black} \textbar \color{Red}Percentage \color{black} of \color{Maroon}women \color{black} with live birth in the past 5 years \textbar (\color{Tan}2003\color{black}, \color{Green}88\color{black}) (2008-09-01 00:00:00, 92)} & Vanilla &  & Although the proportion of \color{Maroon}women \color{black} who have received \color{JungleGreen}antenatal care from a skilled provider\color{black}. \\ \cline{3-5}
 &  & \color{black}Blueprint & \color{Green}88\color{Red}\%\color{black}. What was the rate of \color{JungleGreen}antenatal care from a skilled provider \color{black} in \color{Tan}2003\color{black}? & \color{black}The proportion of \color{Maroon}women \color{black} with \color{JungleGreen}antenatal care from a skilled provider \color{black} in \color{Tan}2003\color{black}. \\ \hline
\end{tabular}
\caption{Output examples from vanilla and blueprint models, colour-coded to show where relevant information from the table has been used in blueprints and verbalisations.}
\label{tab:output-eg}
\end{table*}

\end{document}